\documentclass[runningheads]{llncs}

\usepackage{graphicx}
\usepackage{amsmath,amssymb} % define this before the line numbering.
\usepackage{color}
\usepackage[width=122mm,left=12mm,paperwidth=146mm,height=193mm,top=12mm,paperheight=217mm]{geometry}

\usepackage{hyperref}
\usepackage{sidecap}
\usepackage{booktabs}
\usepackage{lib}
\usepackage{tabularx}
\usepackage{colortbl}
\usepackage{epstopdf}

\newcommand{\rgbd}[0]{RGB-D\xspace}
\newcommand{\arbelaez}[0]{Arbel\'{a}ez\xspace}
\newcommand{\rgb}[0]{RGB\xspace}
\newcommand{\bdo}[0]{B3DO\xspace}
\newcommand{\nyu}[0]{NYUD2\xspace}

\newcommand{\regionAP}[0]{$AP^r$\xspace}
\newcommand{\boxAP}[0]{$AP^b$\xspace}
\newcommand{\rcnn}[0]{R-CNN\xspace}
\newcommand{\cnn}[0]{CNN\xspace}
\newcommand{\maxF}[0]{$\textrm{F}_{\max}$\xspace}
\newcommand{\hha}[0]{HHA\xspace}

\newcommand{\bb}[1]{\textbf{#1}}
\newcommand{\notextbf}[1]{#1}
\newcommand{\pp}[0]{} %\phz}
\newcommand{\insertA}[2]{
\includegraphics[width=#1\textwidth]{#2}
}

\begin{document}
% \renewcommand\thelinenumber{\color[rgb]{0.2,0.5,0.8}\normalfont\sffamily\scriptsize\arabic{linenumber}\color[rgb]{0,0,0}}
% \renewcommand\makeLineNumber {\hss\thelinenumber\ \hspace{6mm} \rlap{\hskip\textwidth\ \hspace{6.5mm}\thelinenumber}}
% \linenumbers

\pagestyle{headings}
\mainmatter
\title{Learning Rich Features from \rgbd Images for Object Detection and Segmentation}

\author{Saurabh Gupta$^1$ \and Ross Girshick$^1$ \and Pablo \arbelaez$^{1,2}$ \and Jitendra Malik$^1$
{\tt\small \{sgupta, rbg, arbelaez, malik\}@eecs.berkeley.edu}}
\institute{$^1$University of California, Berkeley, $^2$Universidad de los Andes, Colombia} 

\titlerunning{Learning Rich Features from \rgbd Images for Detection and Segmentation}
\authorrunning{Saurabh Gupta \and Ross Girshick \and Pablo Arbel\'{a}ez \and Jitendra Malik}

\maketitle

\begin{abstract}
In this paper we study the problem of object detection for \rgbd images using semantically rich image and depth features. We propose a new geocentric embedding for depth images that encodes height above ground and angle with gravity for each pixel in addition to the horizontal disparity. We demonstrate that this geocentric embedding works better than using raw depth images for learning feature representations with convolutional neural networks. Our final object detection system achieves an average precision of 37.3\%, which is a 56\% relative improvement over existing methods. 
We then focus on the task of instance segmentation where we label pixels belonging to object instances found by our detector. For this task, we propose a decision forest approach that classifies pixels in the detection window as foreground or background using a family of unary and binary tests that query shape and geocentric pose features. Finally, we use the output from our object detectors in an existing superpixel classification framework for semantic scene segmentation and achieve a 24\% relative improvement over current state-of-the-art for the object categories that we study. We believe advances such as those represented in this paper will facilitate the use of perception in fields like robotics. 
\end{abstract}

\section{Introduction}

We have designed and implemented an integrated system (\figref{figure1}) for scene understanding from  \rgbd images. The overall architecture is a generalization of the current state-of-the-art system for object detection in RGB images, R-CNN \cite{girshickCVPR14}, where we design each module to make effective use of the additional signal in RGB-D images, namely pixel-wise depth. We go beyond object detection by providing pixel-level support maps for individual objects, such as tables and chairs, as well as a pixel-level labeling of scene surfaces, such as walls and floors. Thus our system subsumes the traditionally distinct problems of object detection and semantic segmentation. 
Our approach is summarized below (source code is available at \url{http://www.cs.berkeley.edu/~sgupta/eccv14/}).
\begin{figure}[t!]
\insertA{1.0}{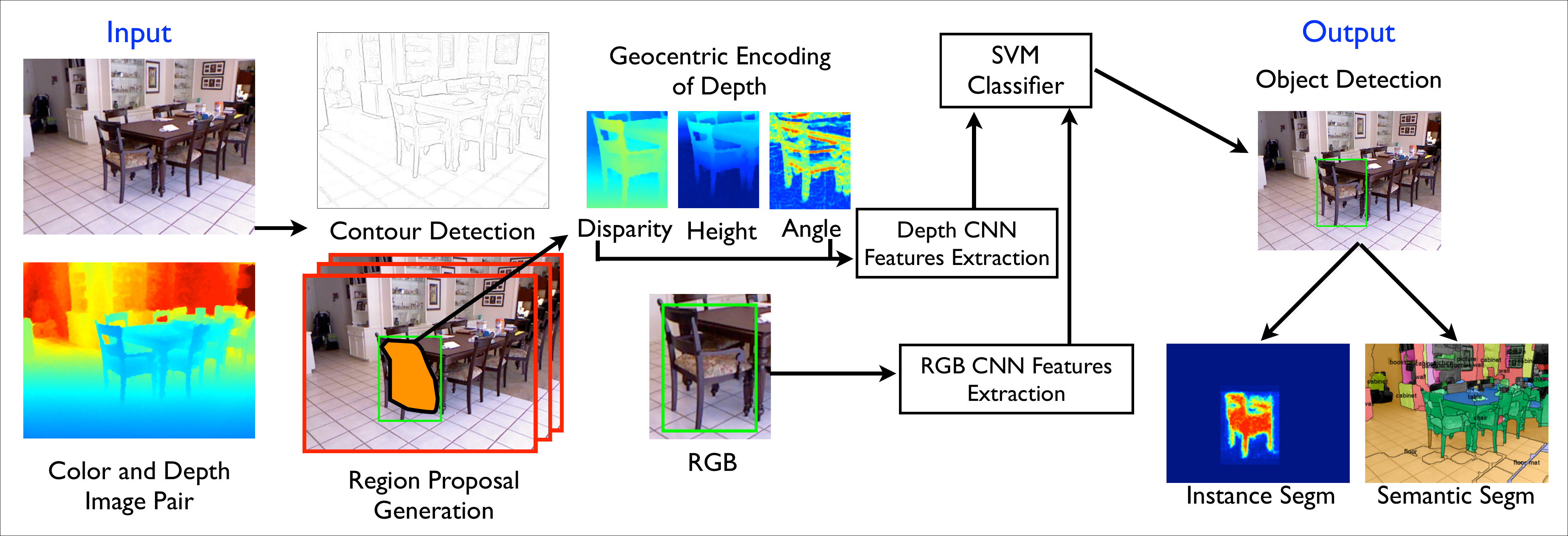}
% \vspace{-1em}
\caption{\textbf{Overview:} from an \rgb and depth image pair, our system detects contours, generates 2.5D region proposals, classifies them into object categories, and then infers segmentation masks for instances of ``thing''-like objects, as well as labels for pixels belonging to ``stuff''-like categories.}
\figlabel{figure1}
% \vspace{-1em}
\end{figure}

\smallskip
\noindent \textbf{\rgbd contour detection and 2.5D region proposals:} RGB-D images enable one to compute depth and normal gradients \cite{guptaCVPR13}, which we combine with the structured learning approach in \cite{dollarICCV13} to yield significantly improved contours. We then use these \rgbd contours to obtain 2.5D region candidates by computing features on the depth and color image for use in the Multiscale Combinatorial Grouping (MCG) framework of \arbelaez \etal \cite{arbelaezCVPR14}.
This module is state-of-the-art for RGB-D proposal generation.

\smallskip
\noindent \textbf{\rgbd object detection:}  Convolutional neural networks ({\cnn}s) trained on \rgb images are the state-of-the-art for detection and segmentation \cite{girshickCVPR14}.
We show that a large \cnn pre-trained on \rgb images can be adapted to generate rich features for depth images.
We propose to represent the depth image by three channels (horizontal disparity, height above ground, and angle with gravity) and show that this representation allows the \cnn to learn stronger features than by using disparity (or depth) alone.
We use these features, computed on our 2.5D region candidates, in a modified \rcnn framework to obtain a 56\% relative improvement in \rgbd object detection, compared to existing methods.

\smallskip
\noindent \textbf{Instance segmentation:} In addition to bounding-box object detection, we also infer pixel-level object masks. We frame this as a foreground labeling task and show improvements over baseline methods.

\smallskip
\noindent \textbf{Semantic segmentation:} Finally, we improve semantic segmentation performance (the task of labeling all pixels with a category, but not differentiating between instances) by using object detections to compute additional features for superpixels in the semantic segmentation system we proposed in \cite{guptaCVPR13}. This approach obtains state-of-the-art results for that task, as well.

\subsection{Related Work}
Most prior work on \rgbd perception has focussed on semantic segmentation \cite{o2p-d,guptaCVPR13,koppulaNIPS11,renCVPR12,silbermanECCV12}, \ie the task of assigning a category label to each pixel. While this is an interesting problem, many practical applications require a richer understanding of the scene. Notably, the notion of an object instance is missing from such an output. Object detection in \rgbd images \cite{b3do,kimCVPR13,laiICRA11,tangACCV13,edmund}, in contrast, focusses on instances, but the typical output is a bounding box. As Hariharan \etal \cite{hariharanECCV14} observe, neither of these tasks produces a compelling output representation. It is not enough for a robot to know that there is a mass of `bottle' pixels in the image. Likewise, a roughly localized bounding box of an individual bottle may be too imprecise for the robot to grasp it. Thus, we propose a framework for solving the problem of instance segmentation (delineating pixels on the object corresponding to each detection) as proposed by \cite{hariharanECCV14,tigheCVPR14}.

Recently, convolutional neural networks \cite{lecun-89e} were shown to be useful for standard \rgb vision tasks like image classification \cite{krizhevskyNIPS12}, object detection \cite{girshickCVPR14}, semantic segmentation \cite{farabetPAMI13} and fine-grained classification \cite{donahueDecaf}.
Naturally, recent works on \rgbd perception have considered neural networks for learning representations from depth images \cite{boISER12,couprieICLR13,socherNIPS12}. Couprie \etal \cite{couprieICLR13} adapt the multiscale semantic segmentation system of Farabet \etal \cite{farabetPAMI13} by operating directly on four-channel \rgbd images from the \nyu dataset. Socher \etal \cite{socherNIPS12} and Bo \etal \cite{boISER12} look at object detection in \rgbd images, but detect small prop-like objects imaged in controlled lab settings. In this work, we tackle uncontrolled, cluttered environments as in the \nyu dataset. More critically, rather than using the \rgbd image directly, we introduce a new encoding that captures the geocentric pose of pixels in the image, and show that it yields a substantial improvement over naive use of the depth channel.

\section{2.5D Region Proposals}
\seclabel{region-proposals}
In this section, we describe how to extend multiscale combinatorial grouping (MCG) \cite{arbelaezCVPR14} to effectively utilize depth cues to obtain 2.5D region proposals.

\subsection{Contour Detection}
\seclabel{contour-detection}
\rgbd contour detection is a well-studied task \cite{dollarICCV13,guptaCVPR13,renNIPS12,silbermanECCV12}. 
Here we combine ideas from two leading approaches, \cite{dollarICCV13} and our past work in \cite{guptaCVPR13}.

In \cite{guptaCVPR13}, we used \emph{gPb-ucm} \cite{arbelaezPAMI11} and proposed local geometric gradients dubbed $NG_{-}$, $NG_{+}$, and $DG$ to capture convex, concave normal gradients and depth gradients. In \cite{dollarICCV13}, Doll\'{a}r \etal proposed a novel learning approach based on structured random forests to directly classify a pixel as being a contour pixel or not.
Their approach treats the depth information as another image, rather than encoding it in terms of geocentric quantities, like $NG_{-}$. While the two methods perform comparably on the \nyu contour detection task (maximum F-measure point in the red and the blue curves in \figref{contour-detection}), there are differences in the the type of contours that either approach produces. \cite{dollarICCV13} produces better localized contours that capture fine details, but tends to miss normal discontinuities that \cite{guptaCVPR13} easily finds (for example, consider the contours between the walls and the ceiling in left part of the image \figref{contour-detection-example}).
We propose a synthesis of the two approaches that combines features from \cite{guptaCVPR13} with the learning framework from \cite{dollarICCV13}.
Specifically, we add the following features.

\smallskip
\noindent \textbf{Normal Gradients:} We compute normal gradients at two scales (corresponding to fitting a local plane in a half-disk of radius 3 and 5 pixels), and use these as additional gradient maps.

\smallskip
\noindent \textbf{Geocentric Pose:} We compute a per pixel height above ground and angle with gravity (using the algorithms we proposed in \cite{guptaCVPR13}. 
These features allow the decision trees to exploit additional regularities, for example that the brightness edges on the floor are not as important as brightness edges elsewhere.

\smallskip
\noindent \textbf{Richer Appearance:} We observe that the \nyu dataset has limited appearance variation (since it only contains images of indoor scenes). To make the model generalize better, we add the soft edge map produced by running the \rgb edge detector of \cite{dollarICCV13} (which is trained on BSDS) on the \rgb image.

\subsection{Candidate Ranking}
\seclabel{region-reranking}
From the improved contour signal, we obtain object proposals by generalizing MCG to \rgbd images.
MCG for \rgb images \cite{arbelaezCVPR14} uses simple features based on the color image and the region shape to train a random forest regressors to rank the object proposals.
We follow the same paradigm, but propose additional geometric features computed on the depth image within each proposal.
We compute: (1) the mean and standard deviation of the disparity, height above ground, angle with gravity, and world $(X,Y,Z)$ coordinates of the points in the region; (2) the region's $(X,Y,Z)$ extent; (3) the region's minimum and maximum height above ground; (4) the fraction of pixels on vertical surfaces, surfaces facing up, and surfaces facing down; (5) the minimum and maximum standard deviation along a direction in the top view of the room. We obtain 29 geometric features for each region in addition to the 14 from the 2D region shape and color image already computed in \cite{arbelaezCVPR14}. Note that the computation of these features for a region decomposes over superpixels and can be done efficiently by first computing the first and second order moments on the superpixels and then combining them appropriately.

\subsection{Results}
\seclabel{region-reranking-results}
We now present results for contour detection and candidate ranking. We work with the \nyu dataset and use the standard split of 795 training images and 654 testing images (we further divide the 795 images into a training set of 381 images and a validation set of 414 images). These splits are carefully selected such that images from the same scene are only in one of these sets.

\smallskip
\noindent \textbf{Contour detection:}
To measure performance on the contour detection task, we plot the precision-recall curve on contours in \figref{contour-detection} and report the standard maximum F-measure metric (\maxF) in \tableref{contour-detection}. We start by comparing the performance of \cite{guptaCVPR13} (Gupta \etal CVPR [RGBD]) and Doll\'{a}r \etal (SE [RGBD]) \cite{dollarICCV13}. We see that both these contour detectors perform comparably in terms of \maxF. \cite{guptaCVPR13} obtains better precision at lower recalls while \cite{dollarICCV13} obtains better precision in the high recall regime. We also include a qualitative visualization of the contours to understand the differences in the nature of the contours produced by the two approaches (\figref{contour-detection-example}). 

Switching to the effect of our proposed contour detector, we observe that adding normal gradients consistently improves precision for all recall levels and \maxF increases by 1.2\% points (\tableref{contour-detection}). 
The addition of geocentric pose features and appearance features improves \maxF by another 0.6\% points, making our final system better than the current state-of-the-art methods by 1.5\% points.\footnote{Doll\'{a}r \etal \cite{dollarPAMI14} recently introduced an extension of their algorithm and report performance improvements (SE+SH[RGBD] dashed red curve in \figref{contour-detection}). We can also use our cues with \cite{dollarPAMI14}, and observe an analogous improvement in performance (Our(SE+SH + all cues) [RGBD] dashed blue curve in \figref{contour-detection}). For the rest of the paper we use the Our(SE+all cues)[RGBD] version of our contour detector.}

\begin{figure}[t!]
\begin{minipage}[b]{0.45\linewidth}
\setlength{\tabcolsep}{-2pt}
\begin{tabular*}{\textwidth}{ccc}
\centering \insertA{0.45}{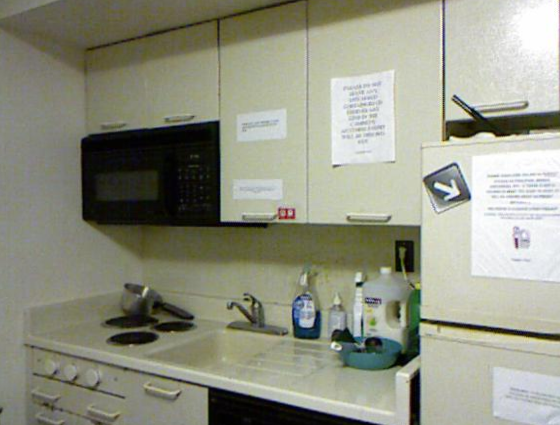} & %\hspace{-1pt}
\centering \insertA{0.45}{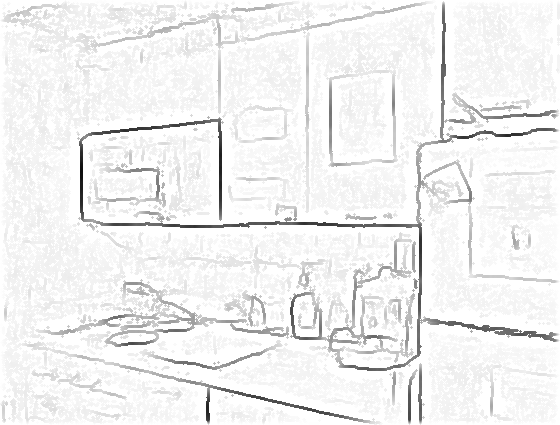} & \\
\centering \insertA{0.45}{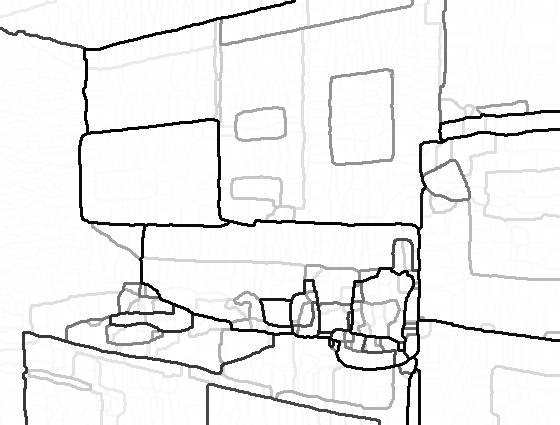} & %\hspace{-1pt}
\centering \insertA{0.45}{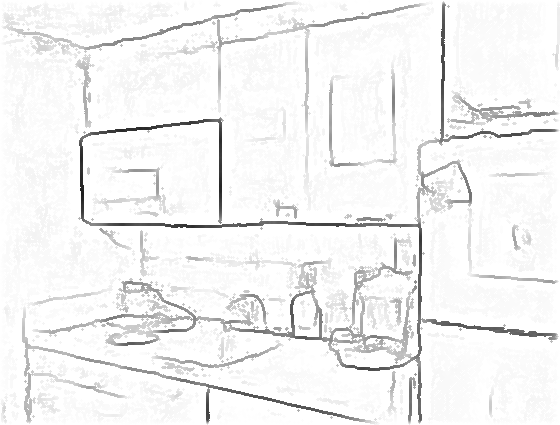} & \\
\end{tabular*}
\caption{\textbf{Qualitative comparison of contours}: Top row: color image, contours from \cite{dollarICCV13}, bottom row: contours from \cite{guptaCVPR13} and contours from our proposed contour detector.} 
\figlabel{contour-detection-example}
\end{minipage}
\hspace{0.5cm}
\begin{minipage}[b]{0.50\linewidth}
\begin{center}
\insertA{0.80}{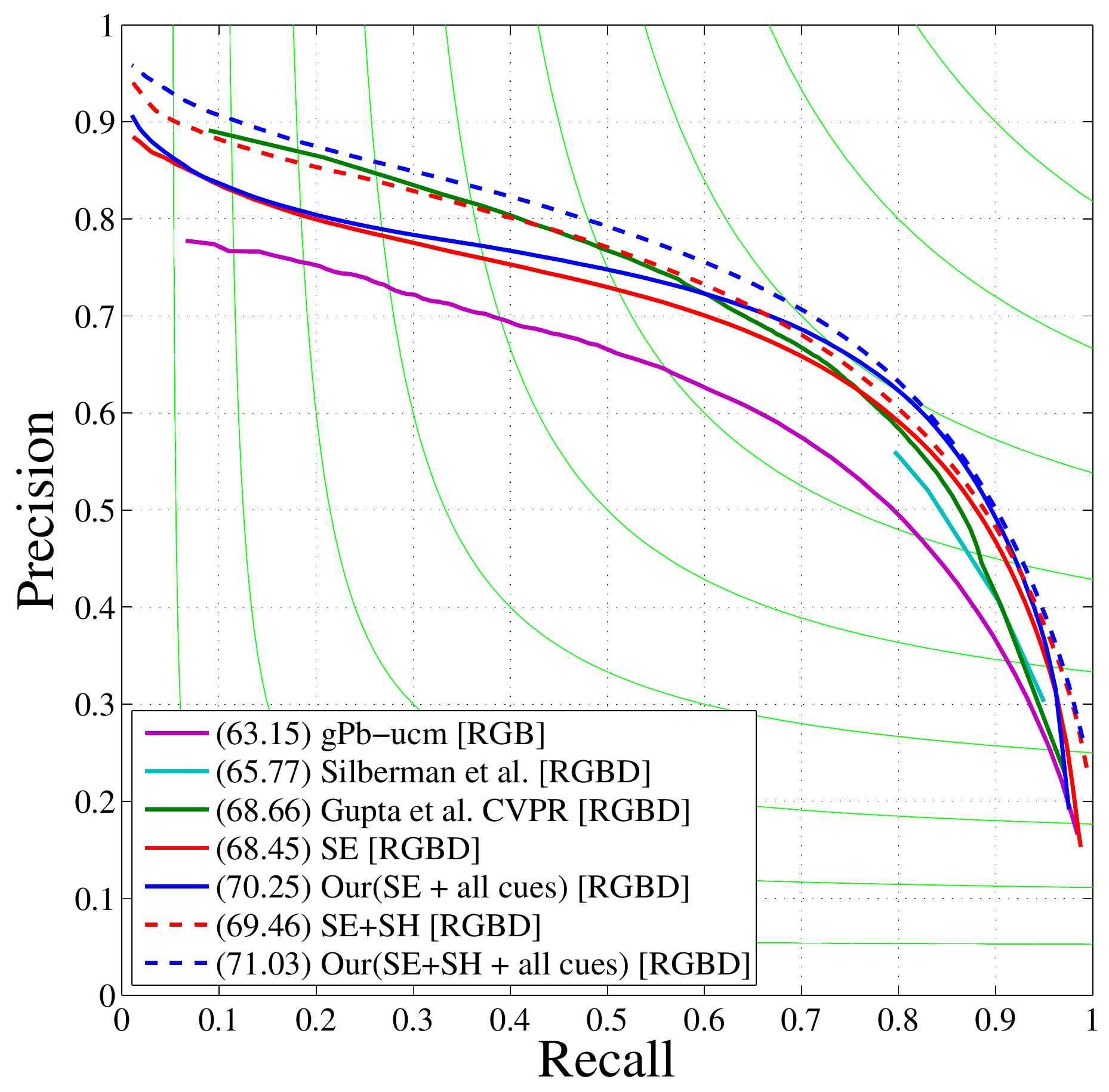}
\caption{\textbf{Precision-recall curve on boundaries on the \nyu dataset.}}
\figlabel{contour-detection}
\end{center}
\end{minipage}
% \vspace{-2em}
\end{figure}

% \begin{wraptable}{r}{7cm}
%\begin{SCtable}
\begin{table}[t!]
\caption{Segmentation benchmarks on \nyu. All numbers are percentages.} 
\setlength{\tabcolsep}{3pt}
\begin{center}
\scalebox{0.75}{
\begin{tabular}{l|l|ccc}
& & ODS (\maxF)       & OIS (\maxF)       & AP \\ \hline
  gPb-ucm & \rgb                                       & 63.15 & 66.12 & 56.20 \\
  Silberman et al. \cite{silbermanECCV12} & \rgbd      & 65.77 & 66.06 & - \\
  Gupta et al. CVPR \cite{guptaCVPR13} & \rgbd         & 68.66 & 71.57 & 62.91 \\
  SE \cite{dollarICCV13} & \rgbd                       & 68.45 & 69.92 & 67.93 \\
  Our(SE + normal gradients) & \rgbd                   & 69.55 & 70.89 & 69.32 \\
  Our(SE + all cues) & \rgbd                           & 70.25 & 71.59 & 69.28 \\ \hline
  SE+SH \cite{dollarPAMI14} & \rgbd                    & 69.46 & 70.84 & 71.88 \\
  Our(SE+SH + all cues) & \rgbd                        & 71.03 & 72.33 & 73.81 \\
\end{tabular}}
\tablelabel{contour-detection}
\end{center}
%\end{SCtable}
% \vspace{-2em}
\end{table}

\begin{SCfigure}
\insertA{0.23}{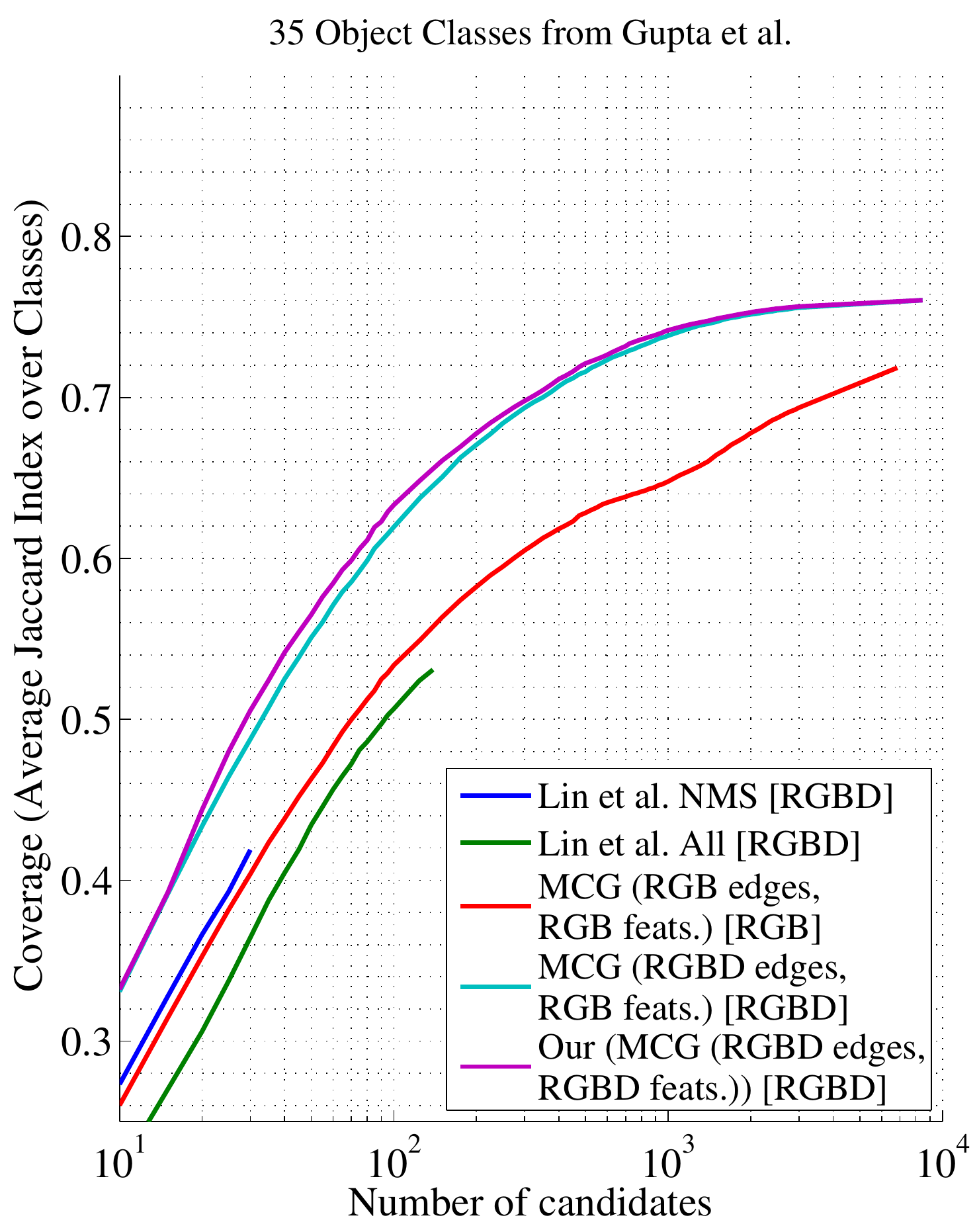}
\insertA{0.23}{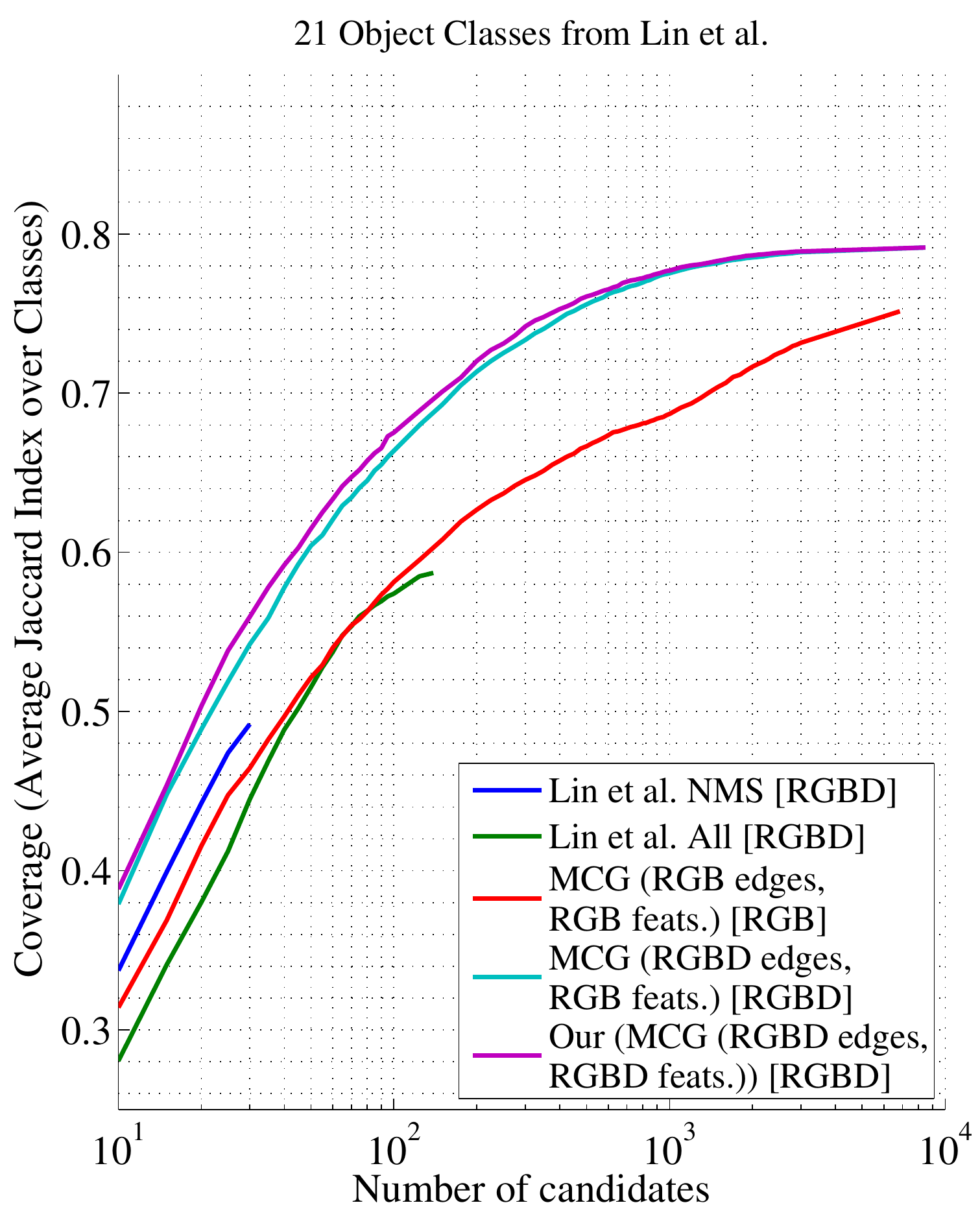}
\caption{\textbf{Region Proposal Quality}: Coverage as a function of the number of region proposal per image for 2 sets of categories: ones which we study in this paper, and the ones studied by Lin~\etal~\cite{linICCV13}. Our depth based region proposals using our improved \rgbd contours work better than Lin \etal's \cite{linICCV13}, while at the same time being more general. Note that the $X$-axis is on a $log$ scale.}
\figlabel{region-ranking}
\end{SCfigure}

\smallskip
\noindent \textbf{Candidate ranking:}
The goal of the region generation step is to propose a pool of candidates for downstream processing (\eg, object detection and segmentation). Thus, we look at the standard metric of measuring the coverage of ground truth regions as a function of the number of region proposals. Since we are generating region proposals for the task of object detection, where each class is equally important, we measure coverage for $K$ region candidates by 
\begin{eqnarray}
\textrm{coverage}(K) &=& \frac{1}{C}\sum_{i = 1}^{C}\left( \frac{1}{N_i} \left( \sum_{j = 1}^{N_i} \max_{k \in [1 \ldots K ]} O\left(R^{l\left(i,j\right)}_k, I^i_j\right) \right) \right),
\end{eqnarray}
where $C$ is the number of classes, $N_i$ is the number of instances for class $i$, $O(a,b)$ is the intersection over union between regions $a$ and $b$, $I^i_j$ is the region corresponding to the $j^{th}$ instance of class $i$, $l\left(i,j\right)$ is the image which contains the $j^{th}$ instance of class $i$, and $R^l_k$ is the $k^{th}$ ranked region in image $l$.

We plot the function $\textrm{coverage}(K)$ in \figref{region-ranking} (left) for our final method, which uses our \rgbd contour detector and \rgbd features for region ranking (black).
As baselines, we show regions from the recent work of Lin \etal \cite{linICCV13} with and without non-maximum suppression, MCG with \rgb contours and \rgb features, MCG with \rgbd contours but \rgb features and finally our system which is MCG with \rgbd contours and \rgbd features. We note that there is a large improvement in region quality when switching from \rgb contours to \rgbd contours, and a small but consistent improvement from adding our proposed depth features for candidate region re-ranking.

Since Lin \etal worked with a different set of categories, we also compare on the subset used in their work (in \figref{region-ranking} (right)).
Their method was trained specifically to return candidates for these classes.
Our method, in contrast, is trained to return candidates for generic objects and therefore ``wastes'' candidates trying to cover categories that do not contribute to performance on any fixed subset.
Nevertheless, our method consistently outperforms \cite{linICCV13}, which highlights the effectiveness and generality of our region proposals.

\section{\rgbd Object Detectors}
\seclabel{detection}
We generalize the \rcnn system introduced by Girshick \etal \cite{girshickCVPR14} to leverage depth information. At test time, R-CNN starts with a set of bounding box proposals from an image, computes features on each proposal using a convolutional neural network, and classifies each proposal as being the target object class or not with a linear SVM.
The CNN is trained in two stages: first, pretraining it on a large set of labeled images with an image classification objective, and then finetuning it on a much smaller detection dataset with a detection objective.

We generalize R-CNN to \rgbd images and explore the scientific question: Can we learn rich representations from depth images in a manner similar to those that have been proposed and demonstrated to work well for \rgb images? 

\subsection{Encoding Depth Images for Feature Learning}

Given a depth image, how should it be encoded for use in a \cnn?
Should the \cnn work directly on the raw depth map or are there transformations of the input that the \cnn to learn from more effectively?

We propose to encode the depth image with three channels at each pixel: horizontal disparity, height above ground, and the angle the pixel's local surface normal makes with the inferred gravity direction.
We refer to this encoding as \hha.
The latter two channels are computed using the algorithms proposed in \cite{guptaCVPR13} 
and all channels are linearly scaled to map observed values across the training dataset to the 0 to 255 range.

The \hha representation encodes properties of geocentric pose that emphasize complementary discontinuities in the image (depth, surface normal and height).
Furthermore, it is unlikely that a \cnn would automatically learn to compute these properties directly from a depth image, especially when very limited training data is available, as is the case with the \nyu dataset.

We use the \cnn architecture proposed by Krizhevsky \etal in \cite{krizhevskyNIPS12} and used by Girshick \etal in \cite{girshickCVPR14}. The network has about 60 million parameters and was trained on approximately 1.2 million \rgb images from the 2012 ImageNet Challenge \cite{ILSVRC12}. We refer the reader to \cite{krizhevskyNIPS12} for details about the network. Our hypothesis, to be borne out in experiments, is that there is enough common structure between our \hha geocentric images and \rgb images that a network designed for \rgb images can also learn a suitable representation for \hha images. As an example, edges in the disparity and angle with gravity direction images correspond to interesting object boundaries (internal or external shape boundaries), similar to ones one gets in \rgb images (but probably much cleaner). 

\smallskip
\noindent \textbf{Augmentation with synthetic data:} 
An important observation is the amount of supervised training data that we have in the \nyu dataset is about one order of magnitude smaller than what is there for PASCAL VOC dataset (400 images as compared to 2500 images for PASCAL VOC 2007). To address this issue, we generate more data for training and finetuning the network. There are multiple ways of doing this: mesh the already available scenes and render the scenes from novel view points, use data from nearby video frames available in the dataset by flowing annotations using optical flow, use full 3D synthetic CAD objects models available over the Internet and render them into scenes. Meshing the point clouds may be too noisy and nearby frames from the video sequence maybe too similar and thus not very useful. Hence, we followed the third alternative and rendered the 3D annotations for \nyu available from \cite{guoICCV13} to generate synthetic scenes from various viewpoints. We also simulated the Kinect quantization model in generating this data (rendered depth images are converted to quantized disparity images and low resolution white noise was added to the disparity values).

\subsection{Experiments}
\seclabel{detection-experiments}
We work with the \nyu dataset and use the standard dataset splits into \emph{train}, \emph{val}, and \emph{test} as described in \secref{region-reranking-results}. The dataset comes with semantic segmentation annotations, which we enclose in a tight box to obtain bounding box annotations. We work with the major furniture categories available in the dataset, such as chair, bed, sofa, table (listed in \tableref{detection-control}).

\smallskip
\noindent \textbf{Experimental setup:} There are two aspects to training our model: finetuning the convolutional neural network for feature learning, and training linear SVMs for object proposal classification. 

\smallskip
\noindent \textbf{Finetuning:} We follow the \rcnn procedure from \cite{girshickCVPR14} using the Caffe \cnn library~\cite{jiaCaffe}.
We start from a \cnn that was pretrained on the much larger ILSVRC 2012 dataset. For finetuning, the learning rate was initialized at $0.001$ and decreased by a factor of 10 every 20k iterations. We finetuned for 30k iterations, which takes about 7 hours on a NVIDIA Titan GPU. 
Following \cite{girshickCVPR14}, we label each training example with the class that has the maximally overlapping ground truth instance, if this overlap is larger than $0.5$, and \emph{background} otherwise. 
All finetuning was done on the \emph{train} set.

\smallskip
\noindent \textbf{SVM Training:} For training the linear SVMs, we compute features either from pooling layer 5 (\emph{pool5}), fully connected layer 6 (\emph{fc6}), or fully connected layer 7 (\emph{fc7}). In SVM training, we fixed the positive examples to be from the ground truth boxes for the target class and the negative examples were defined as boxes having less than $0.3$ intersection over union with the ground truth instances from that class. Training was done on the \emph{train} set with SVM hyper-parameters $C = 0.001$, $B = 10$, $w_1 = 2.0$ using liblinear \cite{liblinear}. We report the performance (detection average precision \boxAP) on the \emph{val} set for the control experiments. For the final experiment we train on \emph{trainval} and report performance in comparison to other methods on the \emph{test} set. At test time, we compute features from the \emph{fc6} layer in the network, apply the linear classifier, and non-maximum suppression to the output, to obtain a set of sparse detections on the test image. 

\renewcommand{\arraystretch}{1.1}
\begin{table}[t!]
\caption{\small{\textbf{Control experiments for object detection on \nyu \emph{val} set}. We investigate a variety of ways to encode the depth image for use in a CNN for feature learning. Results are AP as percentages. See \secref{detection-experiments}}.} 
\tablelabel{detection-control}
\begin{center}
\setlength{\tabcolsep}{1.5pt}
\scalebox{0.75}{
\begin{tabular}{r|cc|cc|cc|c|cc|cc|c}
                & A & B & C & D & E & F & G & H & I & J & K & L \\ \hline
                & DPM & DPM & CNN & CNN & CNN & CNN & CNN & CNN & CNN & CNN & CNN & CNN  \\
finetuned?      &     &     & no  & yes & no  & yes & yes & yes & yes & yes & yes & yes \\
\hline
input channels  & RGB & RGBD & RGB & RGB & disparity & disparity & HHA & HHA & HHA & HHA & HHA & RGB+HHA \\
synthetic data? & & & & & & & & 2x & 15x & 2x & 2x & 2x\\ \hline
\cnn layer      & & & fc6 & fc6 & fc6 & fc6 & fc6 & fc6 & fc6 & pool5 & fc7 & fc6\\ \hline
        bathtub & \phz0.1 & 12.2 & \phz4.9 & \phz5.5 & \phz3.5 & \phz6.1 & 20.4 & 20.7 & 20.7 & 11.1 & 19.9 & \textbf{22.9} \\
            bed & 21.2 & 56.6 & 44.4 & 52.6 & 46.5 & 63.2 & 60.6 & 67.2 & \textbf{67.8} & 61.0 & 62.2 & 66.5 \\
      bookshelf & \phz3.4 & \phz6.3 & 13.8 & 19.5 & 14.2 & 16.3 & 20.7 & 18.6 & 16.5 & 20.6 & 18.1 & \textbf{21.8} \\
            box & \phz0.1 & \phz0.5 & \phz1.3 & \phz1.0 & \phz0.4 & \phz0.4 & \phz0.9 & \phz1.4 & \phz1.0 & \phz1.0 & \phz1.1 & \textbf{\phz3.0} \\
          chair & \phz6.6 & 22.5 & 21.4 & 24.6 & 23.8 & 36.1 & 38.7 & 38.2 & 35.2 & 32.6 & 37.4 & \textbf{40.8} \\
        counter & \phz2.7 & 14.9 & 20.7 & 20.3 & 18.5 & 32.8 & 32.4 & 33.6 & 36.3 & 24.1 & 35.0 & \textbf{37.6} \\
           desk & \phz0.7 & \phz2.3 & \phz2.8 & \phz6.7 & \phz1.8 & \phz3.1 & \phz5.0 & \phz5.1 & \phz7.8 & \phz4.2 & \phz5.4 & \textbf{10.2} \\
           door & \phz1.0 & \phz4.7 & 10.6 & 14.1 & \phz0.9 & \phz2.3 & \phz3.8 & \phz3.7 & \phz3.4 & \phz2.8 & \phz3.3 & \textbf{20.5} \\
        dresser & \phz1.9 & 23.2 & 11.2 & 16.2 & \phz3.7 & \phz5.7 & 18.4 & 18.9 & \textbf{26.3} & 13.1 & 24.7 & 26.2 \\
    garbage-bin & \phz8.0 & 26.6 & 17.4 & 17.8 & \phz2.4 & 12.7 & 26.9 & 29.1 & 16.4 & 21.4 & 25.3 & \textbf{37.6} \\
           lamp & 16.7 & 25.9 & 13.1 & 12.0 & 10.5 & 21.3 & 24.5 & 26.5 & 23.6 & 22.3 & 23.2 & \textbf{29.3} \\
        monitor & 27.4 & 27.6 & 24.8 & 32.6 & \phz0.4 & \phz5.0 & 11.5 & 14.0 & 12.3 & 17.7 & 13.5 & \textbf{43.4} \\
    night-stand & \phz7.9 & 16.5 & \phz9.0 & 18.1 & \phz3.9 & 19.1 & 25.2 & 27.3 & 22.1 & 25.9 & 27.8 & \textbf{39.5} \\
         pillow & \phz2.6 & 21.1 & \phz6.6 & 10.7 & \phz3.8 & 23.4 & 35.0 & 32.2 & 30.7 & 31.1 & 31.2 & \textbf{37.4} \\
           sink & \phz7.9 & \textbf{36.1} & 19.1 & \phz6.8 & 20.0 & 28.5 & 30.2 & 22.7 & 24.9 & 18.9 & 23.0 & 24.2 \\
           sofa & \phz4.3 & 28.4 & 15.5 & 21.6 & \phz7.6 & 17.3 & 36.3 & 37.5 & 39.0 & 30.2 & 34.3 & \textbf{42.8} \\
          table & \phz5.3 & 14.2 & \phz6.9 & 10.0 & 12.0 & 18.0 & 18.8 & 22.0 & 22.6 & 21.0 & 22.8 & \textbf{24.3} \\
     television & 16.2 & 23.5 & 29.1 & 31.6 & 9.7 & 14.7 & 18.4 & 23.4 & 26.3 & 18.9 & 22.9 & \textbf{37.2} \\
         toilet & 25.1 & 48.3 & 39.6 & 52.0 & 31.2 & \textbf{55.7} & 51.4 & 54.2 & 52.6 & 38.4 & 48.8 & 53.0 \\
\hline
           mean & \notextbf{\phz8.4} & \notextbf{21.7} & \notextbf{16.4} & \notextbf{19.7} & \notextbf{11.3} & \notextbf{20.1} & \notextbf{25.2} & \notextbf{26.1} & \notextbf{25.6} & \notextbf{21.9} & \notextbf{25.3} & \textbf{32.5} \\ 
\end{tabular}}
\end{center}
% \vspace{-1em}
\end{table}

\subsection{Results}
\seclabel{detection-results}
We use the PASCAL VOC box detection average precision (denoted as \boxAP following the generalization introduced in \cite{hariharanECCV14}) as the performance metric. Results are presented in \tableref{detection-control}. As a baseline, we report performance of the state-of-the-art non-neural network based detection system, deformable part models (DPM) \cite{lsvm-pami}. First, we trained DPMs on \rgb images, which gives a mean \boxAP of 8.4\% (column A). While quite low, this result agrees with \cite{shrivastavaICCV13}.\footnote{Wang \etal \cite{wangCVPR13} report impressive detection results on NYUD2, however we are unable to compare directly with their method because they use a non-standard train-test split that they have not made available. Their baseline HOG DPM detection results are significantly higher than those reported in \cite{shrivastavaICCV13} and this paper, indicating that the split used in \cite{wangCVPR13} is substantially easier than the standard evaluation split.}
As a stronger baseline, we trained DPMs on features computed from \rgbd images (by using HOG on the disparity image and a histogram of height above ground in each HOG cell in addition to the HOG on the \rgb image). These augmented DPMs (denoted RGBD-DPM) give a mean \boxAP of 21.7\% (column B). We also report results from the method of Girshick \etal \cite{girshickCVPR14}, without and with fine tuning on the \rgb images in the dataset, yielding 16.4\% and 19.7\% respectively (column C and column D). We compare results from layer \emph{fc6} for all our experiments. Features from layers \emph{fc7} and \emph{pool5} generally gave worse performance. 

The first question we ask is: Can a network trained only on \rgb images can do anything when given disparity images? (We replicate each one-channel disparity image three times to match the three-channel filters in the CNN and scaled the input so as to have a distribution similar to \rgb images.)
The \rgb network generalizes surprisingly well and we observe a mean \boxAP of 11.3\% (column E). 
This results confirms our hypothesis that disparity images have a similar structure to \rgb images, and it may not be unreasonable to use an ImageNet-trained CNN as an initialization for finetuning on depth images.
In fact, in our experiments we found that it was always better to finetune from the ImageNet initialization than to train starting with a random initialization.

We then proceed with finetuning this network (starting from the ImageNet initialization), and observe that performance improves to 20.1\% (column F), already becoming comparable to RGBD-DPMs. However, finetuning with our \hha depth image encoding dramatically improves performance (by 25\% relative), yielding a mean \boxAP of 25.2\% (column G). 

We then observe the effect of synthetic data augmentation. Here, we add $2\times$ synthetic data, based on sampling two novel views of the given \nyu scene from the 3D scene annotations made available by \cite{guoICCV13}. We observe an improvement from 25.2\% to 26.1\% mean \boxAP points (column H). However, when we increase the amount of synthetic data further ($15\times$ synthetic data), we see a small drop in performance (column H to I). We attribute the drop to the larger bias that has been introduced by the synthetic data. Guo \etal's \cite{guoICCV13} annotations replace all non-furniture objects with cuboids, changing the statistics of the generated images. More realistic modeling for synthetic scenes is a direction for future research.

We also report performance when using features from other layers: \emph{pool5} (column J) and \emph{fc7} (column K). As expected the performance for \emph{pool5} is lower, but the performance for \emph{fc7} is also lower. We attribute this to over-fitting during finetuning due to the limited amount of data available.

Finally, we combine the features from both the \rgb and the \hha image when finetuned on $2\times$ synthetic data (column L). We see there is consistent improvement from 19.7\% and 26.1\% individually to 32.5\% (column L) mean \boxAP. This is the final version of our system.

We also experimented with other forms of \rgb and D\xspace fusion - early fusion where we passed in a 4 channel \rgbd image for finetuning but were unable to obtain good results (\boxAP of 21.2\%), and late fusion with joint finetuning for \rgb and \hha (\boxAP of 31.9\%) performed comparably to our final system (individual finetuning of \rgb and \hha networks) (\boxAP of 32.5\%). We chose the simpler architecture.

\smallskip
\noindent \textbf{Test set performance:} We ran our final system (column L) on the \emph{test} set, by training on the complete \emph{trainval} set.
Performance is reported in \tableref{detection-test}. We compare against a \rgb DPM, RGBD-DPMs as introduced before. Note that our RGBD-DPMs serve as a strong baseline and are already an absolute 8.2\% better than published results on the \bdo dataset \cite{b3do} (39.4\% as compared to 31.2\% from the approach of Kim \etal \cite{kimCVPR13}, detailed results are in the supplementary material \cite{guptaECCV14Sup}).
We also compare to Lin \etal \cite{linICCV13}. \cite{linICCV13} only produces 8, 15 or 30 detections per image which produce an average $F_1$ measure of 16.60, 17.88 and 18.14 in the 2D detection problem that we are considering as compared to our system which gives an average \maxF measure of 43.70. Precision Recall curves for our detectors along with the 3 points of operation from \cite{linICCV13} are in the supplementary material \cite{guptaECCV14Sup}. 

\smallskip
\noindent \textbf{Result visualizations:} We show some of the top scoring \emph{true positives} and the top scoring \emph{false positives} for our bed, chair, lamp, sofa and toilet detectors in \figref{detection-visualizations}. More figures can be found in the supplementary material \cite{guptaECCV14Sup}.

\newcommand{\insertFIGB}[7]{
  \centering \insertA{#1}{images/#2New/#2-#3-out-small.png} &
  \centering \insertA{#1}{images/#2New/#2-#4-out-small.png} &
  \centering \insertA{#1}{images/#2New/#2-#5-out-small.png} &
  \centering \insertA{#1}{images/#2New/#2-#6-out-small.png} &
  \centering \insertA{#1}{images/#2New/#2-#7-out-small.png} &
}

\begin{figure}[t!]
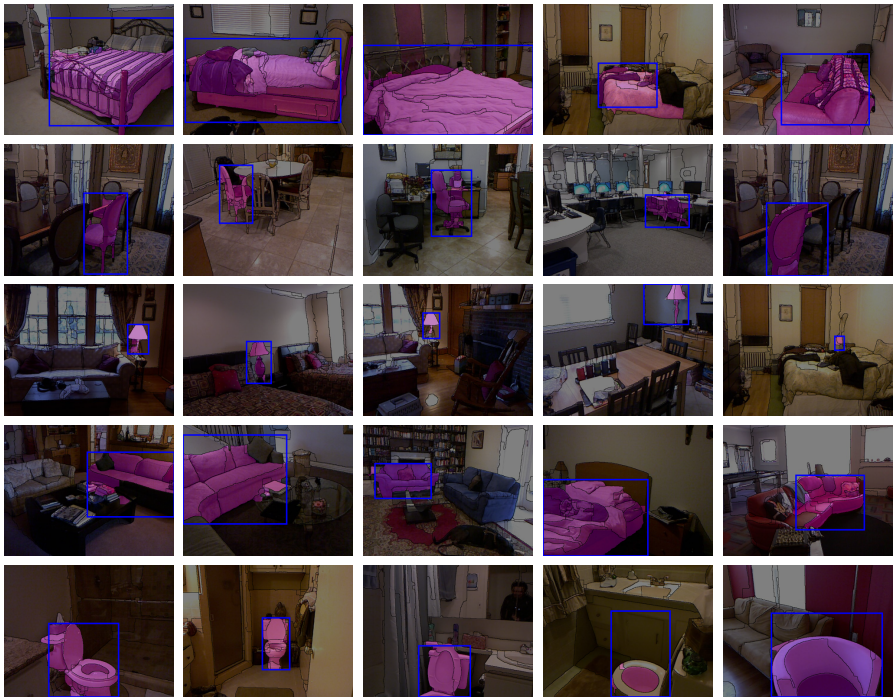

\setlength{\tabcolsep}{-1pt}
\begin{tabular*}{\textwidth}{cccccc}
\insertFIGB{0.185}{bed}{004-tp}{031-tp}{038-tp}{003-fp}{009-fp} \\
\insertFIGB{0.185}{chair}{003-tp}{004-tp}{020-tp}{003-fp}{007-fp} \\
\insertFIGB{0.185}{lamp}{001-tp}{003-tp}{004-tp}{008-fp}{009-fp} \\
\insertFIGB{0.185}{sofa}{003-tp}{005-tp}{010-tp}{006-fp}{008-fp} \\
\insertFIGB{0.185}{toilet}{003-tp}{009-tp}{013-tp}{001-fp}{005-fp} \\
\end{tabular*}
\caption{\textbf{Output of our system}: We visualize some true positives (column one, two and three) and false positives (columns four and five) from our bed, chair, lamp, sofa and toilet object detectors. We also overlay the instance segmentation that we infer for each of our detections. Some of the false positives due to mis-localization are fixed by the instance segmentation.}
\figlabel{detection-visualizations}
% \vspace{-1em}
\end{figure}

\section{Instance Segmentation}
\seclabel{instance-segmentation}
In this section, we study the task of instance segmentation as proposed in~\cite{hariharanECCV14,tigheCVPR14}.
Our goal is to associate a pixel mask to each detection produced by our \rgbd object detector.
We formulate mask prediction as a two-class labeling problem (foreground versus background) on the pixels within each detection window.
Our proposed method classifies each detection window pixel with a random forest classifier and then smoothes the predictions by averaging them over superpixels. 

\subsection{Model Training}
\noindent \textbf{Learning framework:} To train our random forest classifier, we associate each ground truth instance in the \emph{train} set with a detection from our detector. We select the best scoring detection that overlaps the ground truth bounding box by more than 70\%. For each selected detection, we warp the enclosed portion of the associated ground truth mask to a $50 \times 50$ grid. Each of these $2500$ locations (per detection) serves as a training point. 

We could train a single, monolithic classifier to process all $2500$ locations or train a different classifier for each of the $2500$ locations in the warped mask. 
The first option requires a highly non-linear classifier, while the second option suffers from data scarcity. 
We opt for the first option and work with random forests \cite{breiman}, which naturally deal with multi-modal data and have been shown to work well with the set of features we have designed \cite{lim2013sketch,shottonCVPR11}. 
We adapt the open source random forest implementation in \cite{dollarToolbox} to allow training and testing with on-the-fly feature computation. Our forests have ten decision trees.

\smallskip
\noindent \textbf{Features:} We compute a set of feature channels at each pixel in the original image (listed in supplementary material \cite{guptaECCV14Sup}).
For each detection, we crop and warp the feature image to obtain features at each of the $50 \times 50$ detection window locations.
The questions asked by our decision tree split nodes are similar to those in Shotton \etal \cite{shottonCVPR11}, which generalize those originally proposed by Geman \etal \cite{gemanPAMI97}. 
Specifically, we use two question types: \emph{unary questions} obtained by thresholding the value in a channel relative to the location of a point, and \emph{binary questions} obtained by thresholding the difference between two values, at different relative positions, in a particular channel. 
Shotton \etal \cite{shottonCVPR11} scale their offsets by the depth of the point to classify. 
We find that depth scaling is unnecessary after warping each instance to a fixed size and scale.

\smallskip
\noindent \textbf{Testing:} During testing, we work with the top 5000 detections for each category (and 10000 for the chairs category, this gives us enough detections to get to $10\%$ or lower precision). For each detection we compute features and pass them through the random forest to obtain a $50 \times 50$ foreground confidence map. We unwarp these confidence maps back to the original detection window and accumulate the per pixel predictions over superpixels. We select a threshold on the soft mask by optimizing performance on the \emph{val} set.

\subsection{Results}
\seclabel{instance-segmentation-results}
To evaluate instance segmentation performance we use the region detection average precision \regionAP metric (with a threshold of $0.5$) as proposed in~\cite{hariharanECCV14}, which extends the average precision metric used for bounding box detection by replacing bounding box overlap with region overlap (intersection over union).
Note that this metric captures more information than the semantic segmentation metric as it respects the notion of instances, which is a goal of this paper.

We report the performance of our system in \tableref{detection-test}. We compare against three baseline methods: 1) \emph{box} where we simply assume the mask to be the box for the detection and project it to superpixels, 2) \emph{region} where we average the region proposals that resulted in the detected bounding box and project this to superpixels, and 3) \emph{fg mask} where we compute an empirical mask from the set of ground truth masks corresponding to the detection associated with each ground truth instance in the \emph{training} set. We see that \emph{our} approach outperforms all the baselines and we obtain a mean \regionAP of 32.1\% as compared to 28.1\% for the best baseline. The effectiveness of our instance segmentor is further demonstrated by the fact that for some categories the \regionAP is better than \boxAP, indicating that our instance segmentor was able to correct some of the mis-localized detections.

\renewcommand{\arraystretch}{1.1}
\begin{table}[t!]
\caption{\textbf{\emph{Test} set results for detection and instance segmentation on \nyu}: First four rows correspond to box detection average precision, \boxAP, and we compare against three baselines: \rgb DPMs, RGBD-DPMs, and RGB \rcnn. The last four lines correspond to region detection average precision, \regionAP. See \secref{detection-results} and \secref{instance-segmentation-results}.}
\begin{center}
\setlength{\tabcolsep}{1pt}
\scalebox{0.65}{
% \begin{tabular}{r|ccc|cccc}
% \begin{tabular}{c|cccccccccccccccccccc}
\begin{tabular}{c|c|c|c|c|c|c|c|c|c|c|c|c|c|c|c|c|c|c|c|c}
                & mean & bath & bed & book & box & chair & count- & desk & door & dress- & garba- & lamp & monit- & night & pillow & sink & sofa & table & tele & toilet \\ 
                & & tub & & shelf & & & -er & & & -er & -ge bin & & -or & stand & & & & & vision & \\ \hline
  \rgb DPM      & \textbf{\phz9.0} & \phz0.9 & 27.6 & \phz9.0 & \phz0.1 & \phz7.8 & \phz7.3 & \phz0.7 & \phz2.5 & \phz1.4 & \phz6.6 & 22.2 & 10.0 & \phz9.2 & \phz4.3 & \phz5.9 & \phz9.4 & \phz5.5 & \phz5.8 & 34.4 \\
  RGBD-DPM      & \textbf{23.9} & 19.3 & 56.0 & 17.5 & \phz0.6 & 23.5 & 24.0 & \phz6.2 & \phz9.5 & 16.4 & 26.7 & 26.7 & 34.9 & 32.6 & 20.7 & 22.8 & 34.2 & 17.2 & 19.5 & 45.1 \\
\rgb \rcnn      & \textbf{22.5} & 16.9 & 45.3 & 28.5 & \phz0.7 & 25.9 & 30.4 & 9.7 & 16.3 & 18.9 & 15.7 & 27.9 & 32.5 & 17.0 & 11.1 & 16.6 & 29.4 & 12.7 & 27.4 & 44.1 \\
       Our      & \textbf{37.3} & \textbf{44.4} & \textbf{71.0} & \textbf{32.9} & \textbf{\phz1.4} & \textbf{43.3} & \textbf{44.0} & \textbf{15.1} & \textbf{24.5} & \textbf{30.4} & \textbf{39.4} & \textbf{36.5} & \textbf{52.6} & \textbf{40.0} & \textbf{34.8} & \textbf{36.1} & \textbf{53.9} & \textbf{24.4} & \textbf{37.5} & \textbf{46.8} \\ \hline \hline
        box     & \textbf{14.0} & \phz5.9 & 40.0 & \phz4.1 & \phz0.7 & \phz5.5 & \phz0.5 & \phz3.2 & 14.5 & 26.9 & 32.9 & \phz1.2 & 40.2 & 11.1 & \phz6.1 & \phz9.4 & 13.6 & \phz2.6 & 35.1 & 11.9 \\
     region     & \textbf{28.1} & \textbf{32.4} & 54.9 & \phz9.4 & \phz1.1 & 27.0 & 21.4 & \phz8.9 & 20.3 & 29.0 & 37.1 & 26.3 & 48.3 & 38.6 & \textbf{33.1} & 30.9 & 30.5 & 10.2 & 33.7 & 39.9 \\
    fg mask     & \textbf{28.0} & 14.7 & 59.9 & \phz8.9 & \phz1.3 & 29.2 & \phz5.4 & \phz7.2 & 22.6 & 33.2 & 38.1 & 31.2 & \textbf{54.8} & 39.4 & 32.1 & 32.0 & 36.2 & 11.2 & 37.4 & 37.5 \\
        Our     & \textbf{32.1} & 18.9 & \textbf{66.1} & \textbf{10.2} & \textbf{\phz1.5} & \textbf{35.5} & \textbf{32.8} & \textbf{10.2} & \textbf{22.8} & \textbf{33.7} & \textbf{38.3} & \textbf{35.5} & 53.3 & \textbf{42.7} & 31.5 & \textbf{34.4} & \textbf{40.7} & \textbf{14.3} & \textbf{37.4} & \textbf{50.5} \\
\end{tabular}}
\end{center}
\tablelabel{detection-test}
% \vspace{-2em}
\end{table}

\section{Semantic Segmentation}
\seclabel{semantic-segmentation}
\newcolumntype{R}{>{\centering\arraybackslash}X}%
\newcolumntype{T}{>{\columncolor[gray]{0.8}\centering\arraybackslash}X}%

\begin{table}[t!]
\begin{center} 
\caption{\footnotesize{\textbf{Performance on the 40 class semantic segmentation task as proposed by \cite{guptaCVPR13}}: We report the pixel-wise Jaccard index for each of the 40 categories. We compare against 4 baselines: previous approaches from \cite{silbermanECCV12}, \cite{renCVPR12}, \cite{guptaCVPR13} (first three rows), and the approach in \cite{guptaCVPR13} augmented with features from RGBD-DPMs (\cite{guptaCVPR13}+DPM) (fourth row). Our approach obtains the best performance \emph{fwavacc} of 47\%. There is an even larger improvement for the categories for which we added our object detector features, where the average performance \emph{avacc*} goes up from 28.4 to 35.1.
Categories for which we added detectors are shaded in gray (avacc* is the average for categories with detectors).}} 
\tablelabel{ssegm-final}
% \vspace{0.3em}
\scalebox{0.57}{
\begin{tabularx}{1.7\textwidth}{l|RRTTTTTTRTRTRTR}
                & wall & floor & cabinet & bed & chair & sofa & table & door & window & book & picture & counter & blinds & desk & shelves \\
                &  &  &  &  &  &  &  &  &  & shelf & &  &  &  &  \\ \hline
 \cite{silbermanECCV12}-SC & 60.7 & 77.8 & 33.0 & 40.3 & 32.4 & 25.3 & 21.0 & \pp5.9 & 29.7 & \bb{22.7} & 35.7 & 33.1 & 40.6 & \pp4.7 & \pp3.3 \\
 \cite{renCVPR12} & 60.0 & 74.4 & 37.1 & 42.3 & 32.5 & 28.2 & 16.6 & 12.9 & 27.7 & 17.3 & 32.4 & 38.6 & 26.5 & 10.1 & \pp6.1 \\ 
 \cite{guptaCVPR13} & 67.6 & 81.2 & 44.8 & 57.0 & 36.7 & 40.8 & 28.0 & 13.0 & \bb{33.6} & 19.5 & 41.2 & \bb{52.0} & \bb{44.4} & \pp7.1 & \pp4.5 \\ \hline
 \cite{guptaCVPR13}+DPM & 66.4 & \bb{81.5} & 43.2 & 59.4 & 41.1 & 45.6 & \bb{30.3} & 14.2 & 33.2 & 19.6 & \bb{41.5} & 51.8 & 40.7 & \pp6.9 & \bb{\pp9.2} \\
 Ours & \bb{68.0} & 81.3 & \bb{44.9} & \bb{65.0} & \bb{47.9} & \bb{47.9} & 29.9 & \bb{20.3} & 32.6 & 18.1 & 40.3 & 51.3 & 42.0 & \bb{11.3} & \pp3.5 \\
\multicolumn{16}{c}{}
\end{tabularx}
}
\scalebox{0.57}{
\begin{tabularx}{1.7\textwidth}{l|RTTRRRRRRTRRRTR}
                & curtain & dresser & pillow & mirror & floor & clothes & ceiling & books & fridge & tele & paper & towel & shower & box & white \\
                & & & & & mat & & & & & vision & & & curtain & & board \\ \hline
 \cite{silbermanECCV12} & 27.4 & 13.3 & 18.9 & \pp4.4 & \pp7.1 & \pp6.5 & \bb{73.2} & \pp5.5 & \pp1.4 & \pp5.7 & 12.7 & \pp0.1 & \pp3.6 & \pp0.1 & \pp0.0 \\
 \cite{renCVPR12} & 27.6 & \pp7.0 & 19.7 & 17.9 & 20.1 & \bb{9.5} & 53.9 & \bb{14.8} & \pp1.9 & 18.6 & 11.7 & 12.6 & \pp5.4 & \bb{\pp3.3} & \pp0.2 \\ 
 \cite{guptaCVPR13} & 28.6 & 24.3 & 30.3 & 23.1 & 26.8 & \pp7.4 & 61.1 & \pp5.5 & \bb{16.2} & \pp4.8 & 15.1 & \bb{25.9} & \bb{9.7} & \pp2.1 & 11.6 \\ \hline
 \cite{guptaCVPR13}+DPM & 27.9 & 29.6 & \bb{35.0} & \bb{23.4} & \bb{31.2} & \pp7.6 & 61.3 & \pp8.0 & 14.4 & 16.3 & \bb{15.7} & 21.6 & \pp3.9 & \pp1.1 & 11.3 \\
 Ours & \bb{29.1} & \bb{34.8} & 34.4 & 16.4 & 28.0 & \pp4.7 & 60.5 & \pp6.4 & 14.5 & \bb{31.0} & 14.3 & 16.3 & \pp4.2 & \pp2.1 & \bb{14.2} \\

\multicolumn{16}{c}{}
\end{tabularx}
}
\scalebox{0.57}{
\begin{tabularx}{1.7\textwidth}{l|RTTTTTRRRR|RRRRT}
                & person & night & toilet & sink & lamp & bathtub & bag & other & other& other& fwavacc & avacc & mean & pixacc & avacc* \\
                & & stand & & & & & & str & furntr & prop & & & (maxIU) & & \\ \hline
 \cite{silbermanECCV12}-SC & \pp6.6 & \pp6.3 & 26.7 & 25.1 & 15.9 & \pp0.0 & \pp0.0 & \pp6.4 & \pp3.8 & 22.4 & 38.2 & 19.0 & - & 54.6 & 18.4 \\
 \cite{renCVPR12} & \bb{13.6} & \pp9.2 & 35.2 & 28.9 & 14.2 & \pp7.8 & \bb{\pp1.2} & \pp5.7 & \pp5.5 & 9.7 & 37.6 & 20.5 & 21.4 & 49.3 & 21.1 \\
 \cite{guptaCVPR13} & \pp5.0 & 21.5 & 46.5 & 35.7 & 16.3 & 31.1 & \pp0.0 & \pp7.9 & \pp5.7 & 22.7 & 45.2 & 26.4 & 29.1 & 59.1 & 28.4 \\ \hline
 \cite{guptaCVPR13}+DPM & \pp2.2 & 19.9 & 46.5 & \bb{45.0} & 31.3 & 21.5 & \pp0.0 & \bb{\pp9.3} & \pp4.7 & 21.8 & 45.6 & 27.4 & 30.5 & 60.1 & 31.0 \\
 Ours & \pp0.2 & \bb{27.2} & \bb{55.1} & 37.5 & \bb{34.8} & \bb{38.2} & \pp0.2 & \pp7.1 & \bb{\pp6.1} & \bb{23.1} & \bb{47.0} & \bb{28.6} & \bb{31.3} & \bb{60.3} & \bb{35.1} \\
\end{tabularx}
}
\end{center}
% \vspace{-2em}
\end{table}
Semantic segmentation is the problem of labeling an image with the correct category label at each pixel. There are multiple ways to approach this problem, like that of doing a bottom-up segmentation and classifying the resulting superpixels \cite{guptaCVPR13,renCVPR12} or modeling contextual relationships among pixels and superpixels \cite{koppulaNIPS11,silbermanECCV12}. 

Here, we extend our approach from \cite{guptaCVPR13}, which produces state-of-the-art results on this task, and investigate the use of our object detectors in the pipeline of computing features for superpixels to classify them. In particular, we design a set of features on the superpixel, based on the detections of the various categories which overlap with the superpixel, and use them in addition to the features preposed in \cite{guptaCVPR13}.

\subsection{Results}
We report our semantic segmentation performance in \tableref{ssegm-final}. We use the same metrics as \cite{guptaCVPR13}, the frequency weighted average Jaccard Index $fwavacc$\footnote{We calculate the pixel-wise intersection over union for each class independently as in the PASCAL VOC semantic segmentation challenge and then compute an average of these category-wise IoU numbers weighted by the pixel frequency of these categories.}, but also report other metrics namely the average Jaccard Index (\emph{avacc}) and average Jaccard Index for categories for which we added the object detectors (\emph{avacc*}). As a baseline we consider \cite{guptaCVPR13} + DPM, where we replace our detectors with RGBD-DPM detectors as introduced in \secref{detection-results}. We observe that there is an increase in performance by adding features from DPM object detectors over the approach of \cite{guptaCVPR13}, and the \emph{fwavacc} goes up from $45.2$ to $45.6$, and further increase to $47.0$ on adding our detectors. The quality of our detectors is brought out further when we consider the performance on just the categories for which we added object detectors which on average goes up from 28.4\% to 35.1\%.
This 24\% relative improvement is much larger than the boost obtained by adding RGBD-DPM detectors (31.0\% only a 9\% relative improvement over 28.4\%).

\subsubsection{Acknowledgements}: This work was sponsored by ONR SMARTS MURI N00014-09-1-1051, ONR MURI N00014-10-1-0933 and a Berkeley Fellowship. The GPUs used in this research were generously donated by the NVIDIA Corporation. We are also thankful to Bharath Hariharan, for all the useful discussions. We also thank Piotr Doll\'{a}r for helping us with their contour detection code.

\clearpage

\bibliographystyle{splncs03}
\bibliography{refs-rbg}
\end{document}